\documentclass{article}

\usepackage{microtype}
\usepackage{graphicx}
\usepackage{subcaption}
\usepackage{multirow}
\usepackage{algorithm}
\usepackage[T1]{fontenc}
\usepackage{dblfloatfix}
\usepackage{float}  
\usepackage{booktabs}

\usepackage{hyperref}

\usepackage[preprint]{icml2026}

\usepackage{amsmath}
\usepackage{amssymb}
\usepackage{mathtools}
\usepackage{amsthm}

\usepackage[capitalize,noabbrev]{cleveref}

\theoremstyle{plain}
\usepackage{cleveref}
\crefname{figure}{Figure}{Figures}
\crefname{table}{Table}{Tables}
\crefname{table*}{Table}{Tables}
\crefname{appendix}{Appendix}{Appendices}
\crefname{section}{Section}{Sections}
\crefname{equation}{Eq.}{Eqs.}

\theoremstyle{definition}

\theoremstyle{remark}

\usepackage[textsize=tiny]{todonotes}

\icmltitlerunning{BinaryPPO: Efficient Policy Optimization for Binary Classification}

\begin{document}

\twocolumn[
  \icmltitle{BinaryPPO: Efficient Policy Optimization for Binary Classification}

  \icmlsetsymbol{equal}{*}

  \begin{icmlauthorlist}
    \icmlauthor{Punya Syon Pandey}{yyy,comp}
    \icmlauthor{Zhijing Jin}{yyy,comp,sch}
  \end{icmlauthorlist}

  \icmlaffiliation{yyy}{University of Toronto}
  \icmlaffiliation{comp}{Vector Institute}
  \icmlaffiliation{sch}{Max Planck Institute for Intelligent Systems, Tübingen, Germany}

  \icmlcorrespondingauthor{Punya Syon Pandey}{ppandey@cs.toronto.edu}
  \icmlcorrespondingauthor{Zhijing Jin}{zjin@cs.toronto.edu}

  \icmlkeywords{Machine Learning, Binary Classification}
  \vskip 0.3in
]

\printAffiliationsAndNotice{}

\begin{abstract}
Supervised fine-tuning (SFT) is the standard approach for binary classification tasks such as toxicity detection, factuality verification, and causal inference. However, SFT often performs poorly in real-world settings with label noise, class imbalance, or sparse supervision. We introduce \textsc{BinaryPPO}, an offline reinforcement learning large language model (LLM) framework that reformulates binary classification as a reward maximization problem. Our method leverages a variant of Proximal Policy Optimization (PPO) with a confidence-weighted reward function that penalizes uncertain or incorrect predictions, enabling the model to learn robust decision policies from static datasets without online interaction. Across eight domain-specific benchmarks and multiple models with differing architectures, \textsc{BinaryPPO} improves accuracy by 40–60 percentage points, reaching up to 99\%, substantially outperforming supervised baselines. We provide an in-depth analysis of the role of reward shaping, advantage scaling, and policy stability in enabling this improvement. Overall, we demonstrate that confidence-based reward design provides a robust alternative to SFT for binary classification.~\footnote{Our code and additional implementation details are available at \url{https://github.com/psyonp/BinaryPPO}.}
\end{abstract}

\section{Introduction}

Binary classification lies at the core of many critical natural language processing (NLP) tasks, including toxicity detection~\cite{kurita2019robusttoxiccontentclassification, deshpande2023toxicitychatgptanalyzingpersonaassigned}, factuality verification, sentiment analysis~\cite{JIM2024100059, xing-etal-2020-tasty}, and causal inference~\cite{jin2024cladderassessingcausalreasoning}. Despite its simplicity and widespread use, SFT often performs poorly in real-world scenarios where labels are noisy, supervision is sparse, or class distributions are highly imbalanced~\cite{wu2025generalizationsftreinforcementlearning}. These challenges are especially pronounced

\begin{figure}[t]
    \centering
    \includegraphics[width=0.82\linewidth]{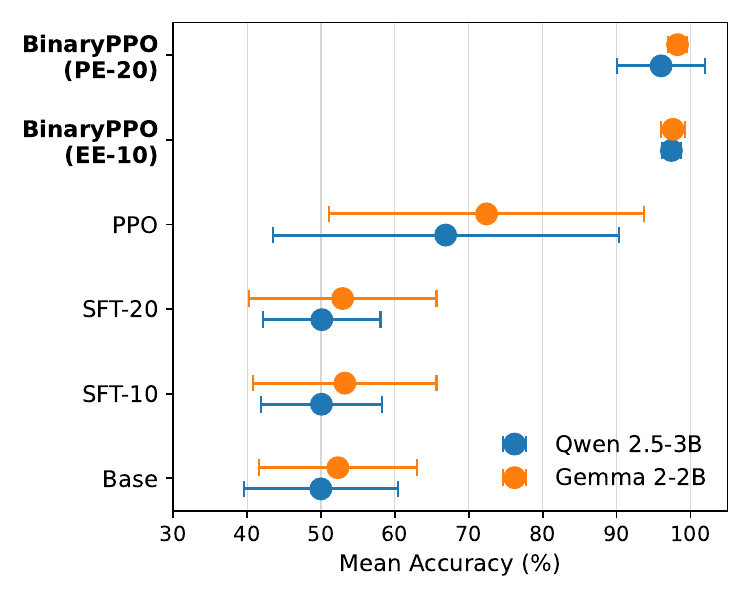}
    \caption{By framing classification as a binary decision problem and optimizing a confidence-based reward function, \textsc{BinaryPPO} achieves consistent performance improvements compared to PPO across interdisciplinary benchmarks.}
    \label{fig:placeholder}
\end{figure}

in AI safety applications, such as harmful content detection or model alignment evaluations~\cite{tong2025badjudgebackdoorvulnerabilitiesllmasajudge, souly2024strongrejectjailbreaks}, where incorrect binary judgments, such as misclassifying a toxic or misleading response, can lead to critical downstream failures~\cite{guo2024biaslargelanguagemodels}. 

Existing supervised methods, even when augmented with techniques like reweighting or noise-robust losses, implicitly assume that labels are reliable ground truth. In practice, however, datasets for tasks like toxicity or factuality often include ambiguous, subjective, or inconsistently annotated examples. Moreover, SFT lacks an explicit mechanism for incorporating model confidence or reasoning under uncertainty. As a result, trained classifiers may overfit to spurious patterns and exhibit brittle behavior when faced with distributional shift ~\cite{luo2025empiricalstudycatastrophicforgetting, kotha2024understandingcatastrophicforgettinglanguage}.

\begin{figure*}[t]
    \centering
    \includegraphics[width=0.87\linewidth]{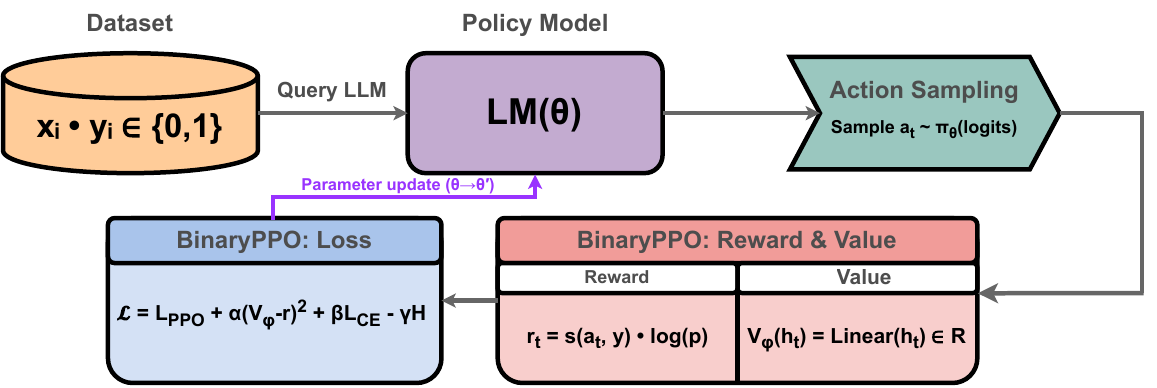}
    \caption{High-level schematic diagram of \textsc{BinaryPPO}. The policy $\pi_\theta$ proposes actions, which are evaluated via confidence-weighted rewards $r(a, y)$ and a value function $V_\phi(h)$; defining the composite loss $\mathcal{L}$ that drives consistent binary performance.}
    \label{fig:placeholder}
\end{figure*}

We propose a fundamentally different approach: \textsc{BinaryPPO}, an offline reinforcement learning (RL) framework that reframes binary classification as a decision-making problem under uncertainty. Rather than minimizing a static loss over potentially flawed labels, \textsc{BinaryPPO} learns a policy to maximize expected reward using Proximal Policy Optimization~\cite{schulman2017proximalpolicyoptimizationalgorithms} with a confidence-weighted reward function. This reward function penalizes incorrect predictions, promoting confident, correct decisions. Because \textsc{BinaryPPO} operates entirely offline, it can be applied to existing datasets without online feedback or interaction, making it practical for sensitive domains like LLM safety evaluation pipelines. Our approach draws on the intuition that classification under noisy supervision is better modeled as a sequential decision problem where an agent must act with imperfect information.

We evaluate \textsc{BinaryPPO} on eight binary classification benchmarks across harmfulness evaluation, sentiment analysis, fact-checking, scientific reasoning, and causal inference. Our method consistently outperforms strong supervised baselines, often yielding tens of percentage points improvement on ambiguous tasks, while maintaining stable loss convergence. In this work, we also analyze the impact of advantage scaling, and policy stability on \textsc{BinaryPPO}.

\section{Related Work}

\paragraph{Intrinsic Reinforcement Learning.} Recent work has demonstrated that RL with sparse or binary signals can substantially improve LLM performance~\cite{zhao2025learningreasonexternalrewards, damani2025binaryrewardstraininglms}. In particular, RL with verifiable rewards (RLVR), has proven effective for reasoning and world modeling, even with extremely limited supervision, including single-example training regimes~\cite{wu2025rlvrworldtrainingworldmodels, wang2025reinforcementlearningreasoninglarge}. Additional efforts have explored replacing external supervision with intrinsic signals, such as model self-certainty, enabling unsupervised learning while matching performance of supervised baselines~\cite{zhao2025learningreasonexternalrewards, laskin2021urlbunsupervisedreinforcementlearning, liu2025honestlanguagemodelsdeductive}. Related alignment methods such as Binary Classifier Optimization~\cite{jung2025binaryclassifieroptimizationlarge} show that binary user feedback can be sufficient to approximate preference-based objectives like Direct Preference Optimization~\cite{rafailov2024directpreferenceoptimizationlanguage}. However, unlike prior work that uses intrinsic signals or optimizes calibration as a separate objective, \textsc{BinaryPPO} integrates model confidence into policy optimization.

\paragraph{Binary Classification.} RL and classification are closely connected, with early work framing classification as a Markov decision processes~\cite{Sutton1998BetweenMA, NIPS2013_0deb1c54, NIPS2004_421b3ac5}, showing competitive performance with standard classifiers~\cite{5967372}. Subsequent theoretical results established direct links between expected cumulative reward and binary classification accuracy, independent of assumptions on state observability~\cite{10.1145/1102351.1102411}. Recent work emphasizes uncertainty and confidence in limited supervision, using agent uncertainty to improve learning from demonstrations~\cite{yippee} and optimizing calibration with correctness in LLM reasoning~\cite{damani2025binaryrewardstraininglms, liu2025llmmicroscopemodelinternals}. At the same time, robustness theory shows that learning reliable binary classifiers can require higher sample complexity under adversarial or distributional shifts~\cite{schmidt2018adversariallyrobustgeneralizationrequires, dowling2024adversarialrobustnessguaranteesquantum}. These results motivate algorithmic designs that stabilize learning under coarse feedback, a gap that \textsc{BinaryPPO} addresses by modifying the policy update dynamics through confidence-aware rewards, rather than relying on additional supervision.

\paragraph{Classifier-Based Alignment.}

A growing body of work argues that causal structure is essential for improving sample efficiency, generalization, and interpretability in RL~\cite{zeng2023surveycausalreinforcementlearning, gasse2021causalreinforcementlearningusing, deng2023causalreinforcementlearningsurvey}. Separately, safety and alignment research for LLMs has increasingly relied on auxiliary classifiers trained to detect undesirable behaviors such as jailbreak attempts, malicious content generation~\cite{sharma2025constitutionalclassifiersdefendinguniversal} and broader surveys of safeguard mechanisms and their limitations~\cite{dong2024safeguardinglargelanguagemodels}. While effective, such approaches typically decouple policy learning from safety enforcement. In contrast, \textsc{BinaryPPO} works directly on policy optimization with binary feedback, providing a complementary alignment approach without separate classifiers or explicit causal models. It applies across domains, including binary causal inference and training safety classifiers.

\section{Problem Formulation}

We consider a binary classification setting where each instance is defined by a natural language input $x \in \mathcal{X}$ and a corresponding label $y \in \{0,1\}$, representing the negative ($0$) and positive ($1$) classes, respectively. Our goal is to train a language model to predict $y$ from $x$ in a reinforcement learning framework that directly aligns probabilistic confidence with task reward.

Let $\mathcal{A} = \{0,1\}$ denote the discrete action space. A stochastic policy $\pi_\theta(a \mid x)$, parameterized by a pretrained language model with parameters $\theta$, defines the probability of taking action $a$ given input $x$:
\setcounter{equation}{0}
\begin{equation}
\pi_\theta(a \mid x) = \Pr(a \mid x; \theta), \quad a \in \mathcal{A}.
\end{equation}

The objective of the policy is to maximize the expected reward
\begin{equation}
J(\theta) = \mathbb{E}_{x \sim \mathcal{D},\, a \sim \pi_\theta} [r(x,a,y)],
\end{equation}
where $r(x,a,y)$ quantifies the alignment between model confidence and ground-truth label consistency.  

\section{Method}
\label{sec:method}

We introduce \textsc{BinaryPPO} (\cref{alg:binaryppo}), a variant of Proximal Policy Optimization~(\cref{eq:ppo}) designed for binary decision tasks, where the reward signal depends on model confidence rather than solely on discrete correctness.
\begin{equation}
\resizebox{0.9\linewidth}{!}{$
\begin{aligned}
\mathcal{L}_{\mathrm{PPO}}(\theta)
&= \mathbb{E}_{x,a \sim \pi_{\theta_{\text{old}}}}
\Bigl[
\min \Bigl(
  \rho_\theta(x,a)\, A(x,a,y), \\
&\qquad
  \mathrm{clip}\bigl(\rho_\theta(x,a), 1-\epsilon, 1+\epsilon\bigr)\,
  A(x,a,y)
\Bigr)
\Bigr]
\end{aligned}
$}
\label{eq:ppo}
\end{equation}

\subsection{Probabilistic Reward Function}

We define a continuous \textit{probabilistic reward}:
\begin{align}
r(x, a, y) &= \kappa \cdot s(a,y) \cdot f\big(\pi_{\theta_{\text{old}}}(a \mid x)\big),
\end{align}
where:
\begin{itemize}
    \item $s(a,y)$ encodes the polarity of correctness, producing a positive signal for correct predictions and a negative signal for incorrect ones
    \item $f(p)$ is a monotonic confidence-dependent shaping function, specifically defined as
\begin{align}
    f(p) = \log(p)
    .
\end{align}
    \item $\kappa$ is a tunable scaling constant controlling reward magnitude
\end{itemize}

\setcounter{table}{1}
\begin{table*}[b]
\centering
\small
\setlength{\tabcolsep}{6pt}
\resizebox{\textwidth}{!}{
\begin{tabular}{l l cccccccc}
\toprule
\addlinespace[0.3em]
\textbf{\textsc{Model}} &
\textbf{\textsc{Method}} &
\textbf{\textsc{CLadder}} &
\textbf{\textsc{SciRIFF}} &
\textbf{\textsc{BoolQ}} &
\textbf{\textsc{Fever}} &
\textbf{\textsc{IMDB}} &
\textbf{\textsc{Mod}} &
\textbf{\textsc{DJ}} &
\textbf{\textsc{JB}} \\
\midrule
\addlinespace[0.3em]
\multirow{6}{*}{\textbf{Qwen 2.5-3B}} 
 & Base    & 49.87 & 37.29 & 46.15 & 50.25 & 54.97 & 37.98 & 72.80 & 50.50 \\
 & SFT (10 Epochs) & 50.23 & 38.43 & 49.27 & 49.88 & 55.15 & 40.06 & 66.47 & 51.00 \\
 & SFT (20 Epochs)  & 49.75 & 39.51 & 48.76 & 49.68 & 55.38 & 42.80 & 67.48 & 47.50 \\
 & PPO  & 50.00 & 76.68 & 37.71 & 39.97 & 51.14 & 97.50 & 97.41 & 84.50 \\
 & \textbf{BinaryPPO (EE-10)}  & 94.97 & 98.74 & 97.35 & 95.57 & 98.26 & 98.39 & 97.99 & 98.00 \\
 & \textbf{BinaryPPO (PE-20)}   & 97.24 & 98.55 & 97.52 & 97.31 & 80.33 & \textbf{99.05} & \textbf{99.10} & \textbf{99.00} \\
\addlinespace
\multirow{6}{*}{\textbf{Gemma 2-2B}}
 & Base    & 49.94 & 47.91 & 54.97 & 49.69 & 51.84 & 35.60 & 76.85 & 51.50 \\
 & SFT (10 Epochs)  & 50.05 & 50.51 & 56.32 & 50.02 & 52.06 & 34.70 & 82.17 & 50.00 \\
 & SFT (10 Epochs)  & 49.91 & 50.44 & 57.30 & 50.01 & 51.48 & 33.27 & 81.99 & 49.00 \\
 & PPO  & 50.05 & 71.62 & 88.68 & 39.97 & 50.00 & 96.25 & 95.13 & 87.50 \\
 & \textbf{BinaryPPO (EE-10)}  & 94.17 & 98.67 & 98.08 & 95.72 & 97.57 & 98.63 & \textbf{99.18} & \textbf{99.00} \\
 & \textbf{BinaryPPO (PE-20)}   & 96.35 & \textbf{99.56} & \textbf{99.05} & \textbf{97.89} & \textbf{99.10} & \textbf{99.46} & 98.51 & 96.00 \\
\bottomrule
\end{tabular}
}
\caption{Accuracy (\%) across benchmarks for Qwen-2.5-3B Instruct and Gemma-2-2B Instruct under different training regimes. Results are reported for the base models, supervised fine-tuning (SFT), and \textsc{BinaryPPO}. EE-10 denotes an equal allocation of 5 epochs each to exploration (probabilistic token selection) and exploitation (greedy token selection), while PE-20 denotes pure exploration for 20 epochs.}
\label{tab:main_results}
\end{table*}

We further introduce a learned value function $V_\phi(x)$ parameterized by $\phi$, which approximates the expected reward under the policy:
\begin{equation}
V_\phi(x) \approx \mathbb{E}_{a \sim \pi_{\theta_{\text{old}}}}[r(x,a,y)].
\end{equation}
The advantage is defined as
\begin{equation}
A(x,a,y) = r(x,a,y) - V_\phi(x),
\end{equation}
and serves as the signal for updating the policy in a direction that improves expected return while maintaining stability via trust-region constraints. 

\subsection{Objective}

\begin{algorithm}[H]
\caption{\textsc{BinaryPPO}}
\label{alg:binaryppo}
\begin{algorithmic}[1]
\REQUIRE Dataset $\mathcal{D} = \{(x_i, y_i)\}_{i=1}^N$, policy $\pi_\theta$, value network $V_\phi$, clip ratio $\epsilon$
\STATE Initialize $\pi_{\theta_{\text{old}}} \gets \pi_\theta$
\FOR{each training epoch}
    \FOR{minibatch $(x,y) \in \mathcal{D}$}
        \STATE Sample action $a \sim \pi_{\theta_{\text{old}}}(\cdot \mid x)$
        \STATE Compute reward $r(x,a,y) = \kappa \, s(a,y) \, f(\pi_{\theta_{\text{old}}}(a \mid x))$
        \STATE Compute advantage $A(x,a,y) = r(x,a,y) - V_\phi(x)$
        \STATE Evaluate $\mathcal{L}_{\mathrm{PPO}}$, $\mathcal{L}_{\mathrm{value}}$, $\mathcal{L}_{\mathrm{CE}}$, and $\mathcal{H}$
        \STATE Update $(\theta,\phi)$ by descending $\mathcal{L}(\theta,\phi)$
    \ENDFOR
    \STATE Set $\pi_{\theta_{\text{old}}} \gets \pi_\theta$
\ENDFOR
\end{algorithmic}
\end{algorithm}

\textsc{BinaryPPO} provides a general framework for confidence-aware policy optimization in binary decision tasks. By parameterizing reward shaping through $f$, the method enables flexible experimentation across binary classification tasks for LLMs.

The total optimization objective balances policy, value, supervised, and entropy terms:
\begin{align}
\mathcal{L}(\theta, \phi) 
&= \mathcal{L}_{\mathrm{PPO}}(\theta) 
+ \alpha \, \mathbb{E}\bigl[(V_\phi(x) - r(x,a,y))^2\bigr] \notag \\
&\quad + \beta \, \mathbb{E}\Bigl[y \log \pi_\theta(1 \mid x) \notag\\
&\qquad + (1-y) \log \pi_\theta(0 \mid x) \Bigr] \notag \\
&\quad - \gamma \, \mathbb{E}\bigl[\mathcal{H}(\pi_\theta(\cdot \mid x))\bigr].
\label{binaryppo_equation}
\end{align}
where $\alpha, \beta, \gamma > 0$ are tunable hyperparameters controlling the trade-off between value regression, supervised regularization, and entropy-driven exploration and $\mathcal{H}(\pi_\theta(\cdot \mid x))$ denotes the Shannon entropy of the policy distribution over actions, defined as
\[
\mathcal{H}(\pi_\theta(\cdot \mid x)) = - \sum_{a} \pi_\theta(a \mid x) \log \pi_\theta(a \mid x), \tag{9}
\]
which encourages exploration by penalizing overconfident (low-entropy) policies.

\begin{figure*}[t]
    \centering
    \begin{subfigure}[b]{0.43\textwidth}
        \centering
        \includegraphics[width=\linewidth]{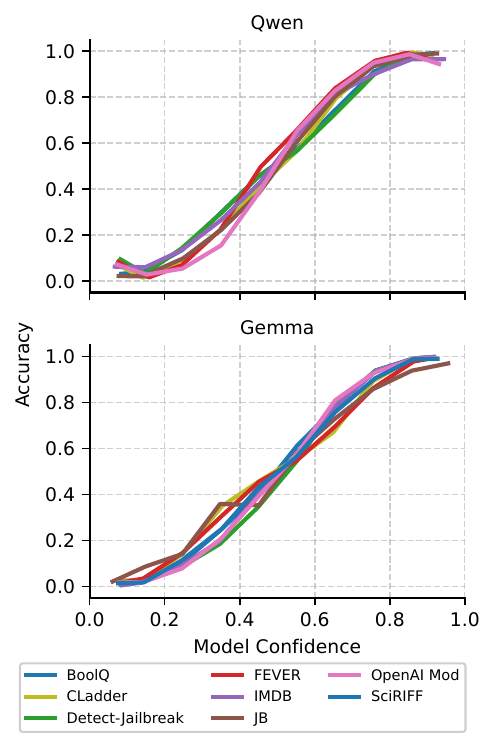}
        \caption{Relationship between post-training accuracy and instilled confidence across Qwen-2.5-3B and Gemma-2-2B models, where increasing confidence aligns with improved accuracy. This indicates alignment between confidence and correctness for \textsc{BinaryPPO}.}
        \label{fig:acc_conf}
    \end{subfigure}
    \hfill
    \begin{subfigure}[b]{0.43\textwidth}
        \centering
        \includegraphics[width=\linewidth]{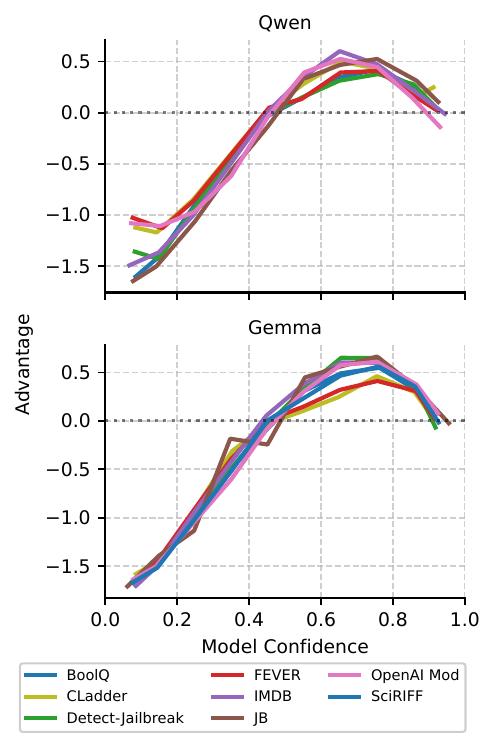}
        \caption{Post-training advantage versus instilled confidence across Qwen-2.5-3B and Gemma-2-2B models, where advantage peaks at intermediate confidence. This reflects maximal corrective signal when predictions are moderately confident for \textsc{BinaryPPO}.}
        \label{fig:adv_conf}
    \end{subfigure}
    \caption{Effect of instilled confidence on post-training behavior under \textsc{BinaryPPO}. 
    (a) Accuracy increases monotonically with confidence, indicating alignment between confidence and correctness. 
    (b) Advantage is maximized at intermediate confidence, suggesting the strongest learning signal occurs for moderately confident predictions (0.6-0.8).}
    \label{fig:combined}
\end{figure*}

\subsection{Technical Contributions}
\textsc{BinaryPPO} introduces key innovations for confidence-aware binary classification. By using a \emph{probabilistic reward function} that scales with model confidence, the framework combines \emph{policy, value, supervised, and entropy optimization} for stable training (\cref{binaryppo_equation}) and implements a flexible \emph{exploration-exploitation schedule}. This balances stochastic exploration with exploitation to enable rapid learning across cross-domain datasets (Tables~\ref{tab:main_results}, \ref{tab:ablations}). 

\setcounter{table}{3}
\begin{table*}[t]
\centering
\small
\setlength{\tabcolsep}{6pt}
\resizebox{\textwidth}{!}{%
\begin{tabular}{l l cccccccc}
\toprule
\addlinespace[0.3em]
\textbf{\textsc{Model}} &
\textbf{\textsc{Ablation}} &
\textbf{\textsc{CLadder}} &
\textbf{\textsc{SciRIFF}} &
\textbf{\textsc{BoolQ}} &
\textbf{\textsc{Fever}} &
\textbf{\textsc{IMDB}} &
\textbf{\textsc{Mod}} &
\textbf{\textsc{DJ}} &
\textbf{\textsc{JB}} \\
\midrule
\addlinespace[0.3em]
\multirow{4}{*}{\textbf{Qwen 2.5-3B}} 
 & Removing Entropy Regularization & 49.87 & 42.48 & 12.46 & 30.77 & 7.74 & 23.33 & 11.28 & 55.00 \\
 & Equal Data Sampling & 60.59 & 92.67 & 91.99 & 92.06 & 77.71 & 97.44 & 79.66 & 60.50 \\
  & \textbf{BinaryPPO (our method)}  & \textbf{94.97} & \textbf{98.74} & \textbf{97.35} & \textbf{95.57} & \textbf{98.26} & \textbf{98.39} & \textbf{97.99} & \textbf{98.00} \\
\addlinespace
\multirow{4}{*}{\textbf{Gemma 2-2B}}
 & Removing Entropy Regularization & 50.37 & 18.20 & 29.80 & 42.55 & 4.82 & 18.93 & 1.96 & 64.50 \\
 & Equal Data Sampling  & 89.00 & 95.45 & 95.57 & 86.00 & 93.19 & 96.55 & 95.21 & 78.50 \\
  & \textbf{BinaryPPO (our method)}  & \textbf{94.17} & \textbf{98.67} & \textbf{98.08} & \textbf{95.72} & \textbf{97.57} & \textbf{98.63} & \textbf{99.18} & \textbf{99.00} \\
\bottomrule
\end{tabular}
}
\caption{Effect of individual training components on \textsc{BinaryPPO} performance. We report accuracy (\%) across benchmarks for Qwen-2.5-3B and Gemma-2-2B under targeted ablations of entropy regularization and equal data sampling, revealing their distinct roles in model stability and generalization compared to the full \textsc{BinaryPPO} approach.}
\label{tab:ablations}
\end{table*}

\section{Experimental Setup}

\textbf{Model Selection.} We evaluate our approach on two open-source base models: Qwen-2.5-3B~\cite{qwen2025qwen25technicalreport} and Gemma-2-2B~\cite{gemmateam2024gemma2improvingopen}. Their open weights also permit SFT and PPO training, necessary for \textsc{BinaryPPO}. Additionally, base models provide a strong baseline for observing performance improvements.

\textbf{Dataset Selection.} We evaluate our method using a set of publicly available datasets covering a range of reasoning and classification tasks. These include the causal reasoning dataset CLadder \cite{jin2024cladderassessingcausalreasoning}, scientific reasoning via SciRIFF \cite{wadden2025sciriffresourceenhancelanguage}, general boolean question-bank via BoolQ \cite{clark2019boolqexploringsurprisingdifficulty}, a 10k-sample subset of the fact-checking FEVER \cite{thorne-etal-2018-fever} dataset, a 25k-sample subset of the movie-review classification IMDB \cite{maas-EtAl:2011:ACL-HLT2011} benchmark, the OpenAI Moderation benchmark~\cite{openai2022moderation}, and two jailbreak detection datasets, Detect-Jailbreak \cite{SCBSZ24, zou2023universal} and JailbreakBench \cite{chao2024jailbreakbench}. We provide additional summary statistics and sample sizes in~\cref{tab:datasets}, with additional descriptions of each dataset in~\cref{ds_description}.

\setcounter{table}{0}
\begin{table}[H]
    \centering
    \small             
    \renewcommand{\arraystretch}{1} 
    \begin{tabular}{lccc}
        \toprule
        \textsc{\textbf{Dataset}} & \textsc{\textbf{Samples}} & \textsc{\textbf{Tokens}} & \textsc{\textbf{Vocabulary}} \\
        \midrule
        \textsc{CLadder} & 10,112 & 1,301,036 & 724 \\
        \textsc{SciRIFF} & 1,582 & 800,168 & 20,178 \\
        \textsc{BoolQ} & 9,427 & 1,284,858 & 37,688 \\
        \textsc{FEVER} & 10,000 & 390,479 & 25,515 \\
        \textsc{IMDB} & 25,000 & 7,461,366 & 47,187 \\
        \textsc{OpenAI} & 1,680 & 260,933 & 24,585 \\
        \textsc{DJ} & 18,258 & 6,631,894 & 68,308 \\
        \textsc{JB} & 200 & 2,856 & 947 \\
        \bottomrule
    \end{tabular}
    \caption{Summary of datasets used in experiments, including prompt sample sizes, token counts, and vocabulary sizes.}
    \label{tab:datasets}
\end{table}

\section{Experiments}

We evaluate \textsc{BinaryPPO} on both in-distribution and out-of-distribution queries. To contextualize its performance, we compare against established SFT and PPO baselines and perform component-specific ablation studies. Beyond final performance, we analyze training dynamics via advantage, confidence, and evolution metrics throughout the \textsc{BinaryPPO} pipeline. We further conduct normative evaluations to assess bias transfer across toxicity and gender bias benchmarks.

\subsection{In-Distribution Performance} 

We first evaluate \textsc{BinaryPPO}'s post-training performance on all datasets, with classification accuracy as the primary metric. Across all in-distribution datasets, \textsc{BinaryPPO} consistently achieves the highest accuracy compared to baseline methods such as vanilla PPO and SFT, reaching performance up to 99.56\% (\cref{tab:main_results}). In addition to improved accuracy, our method yields calibrated confidence estimates over training, driven by an advantage formulation on model confidence (Figures~\ref{fig:acc_conf}, \ref{fig:adv_conf}). In contrast to PPO, \textsc{BinaryPPO} avoids diminishing returns in tasks involving causal identification and fact checking.

\subsection{Out-of-Distribution Performance} 
\label{ood}

\textbf{Causality.} To evaluate generalization in causal reasoning, we perform out-of-distribution (OOD) testing by training \textsc{BinaryPPO} on the original CLadder dataset and evaluating on a reconstructed validation split (\cref{cladder}). As shown in \cref{tab:ood}, the resulting models retain meaningful causal accuracy under distribution shift. In particular, the Qwen 2.5-3B model achieves 50.00\% accuracy on CLadder OOD, while the Gemma 2-2B model substantially improves to 87.46\%, indicating stronger robustness to causal perturbations. These results suggest that \textsc{BinaryPPO} is transferable causal decision boundaries across varying model architectures rather than overfitting to dataset-specific artifacts, consistent with its strong in-domain performance.

\textbf{Harmfulness detection.} We further assess moderation generalization via cross-dataset evaluation between the JB and OpenAI moderation datasets. Specifically, models trained on one dataset are evaluated on the other without additional fine-tuning. As reported in \cref{tab:ood}, \textsc{BinaryPPO} demonstrates consistent cross-dataset performance, with accuracies of 55.42\% (Moderation) and 58.50\% (JB) for Qwen 2.5-3B, and 62.32\% (Moderation) and 53.50\% (JB) for Gemma 2-2B. These scores perform better than base and SFT performance, remaining well above chance and indicate non-trivial generalization across adversarial and stylistically distinct moderation benchmarks. Taken together, these findings suggest that \textsc{BinaryPPO} captures shared signals of harmfulness rather than dataset-specific heuristics.

\setcounter{table}{2}
\begin{table}[H]
    \centering
    \small
    \renewcommand{\arraystretch}{1.1}
    \begin{tabular}{lccc}
        \toprule
        \textsc{\textbf{Model}} & \textsc{\textbf{CLadder}} & \textsc{\textbf{Mod}} & \textsc{\textbf{JB}} \\
        \midrule
        \textsc{Qwen 2.5-3B} & 50.00 & 55.42 & 58.50 \\
        \textsc{Gemma 2-2B} & 87.46 & 62.32 & 53.50 \\
        \bottomrule
    \end{tabular}
    \caption{OOD performance (\%) of \textsc{BinaryPPO}, showing robust causal and moderation generalization across datasets.}
    \label{tab:ood}
\end{table}

\subsection{Ablation Studies}

We conduct ablation studies to isolate the contributions of the core design choices in \textsc{BinaryPPO} (\cref{tab:ablations}). Across both base models, removing either component leads to substantial and systematic performance degradation.

\paragraph{Removing entropy regularization.}
Replacing \textsc{BinaryPPO} with non entropy-regularized PPO results in near-collapse on most tasks. For Qwen~2.5-3B, average accuracy drops from 97.28 to 29.87, with especially severe failures on BoolQ (97.35 $\rightarrow$ 12.46), IMDB (98.26 $\rightarrow$ 7.74), and DJ (97.99 $\rightarrow$ 11.28). Similar trends appear for Gemma~2-2B, where performance degrades from 97.63 to 28.64. This indicates that entropy regularization is essential to stabilize learning in binary-decision settings and in avoiding under-exploration or degenerate policies.

\paragraph{Removing equal data sampling.}
Using equal data sampling without the \textsc{BinaryPPO} objective substantially improves over entropy regularization but still underperforms the full method. On Qwen~2.5-3B, mean accuracy increases to 81.83 but remains 15.5 points below \textsc{BinaryPPO}, with notable gaps on CLadder (60.59 vs.\ 94.97) and JB (60.50 vs.\ 98.00). For Gemma~2-2B, equal sampling achieves a mean of 91.18, yet still trails \textsc{BinaryPPO} by over 6 points. These results suggest that while balanced sampling mitigates class imbalance, it cannot fully address policy instability without the binary-structured objective.

Across all eight benchmarks and both model scales, \textsc{BinaryPPO} consistently dominates its ablated variants, achieving near-saturated performance ($\approx$98--99\%) and reducing variance across tasks. The results demonstrate that \emph{both} the binary-specific PPO formulation and equal data sampling are necessary for stable optimization and high accuracy in tasks involving binary classification via RL procedures.

\subsection{Training Dynamics: Loss-Policy Curves}
\begin{figure}[H]
    \centering
    \includegraphics[width=1\linewidth]{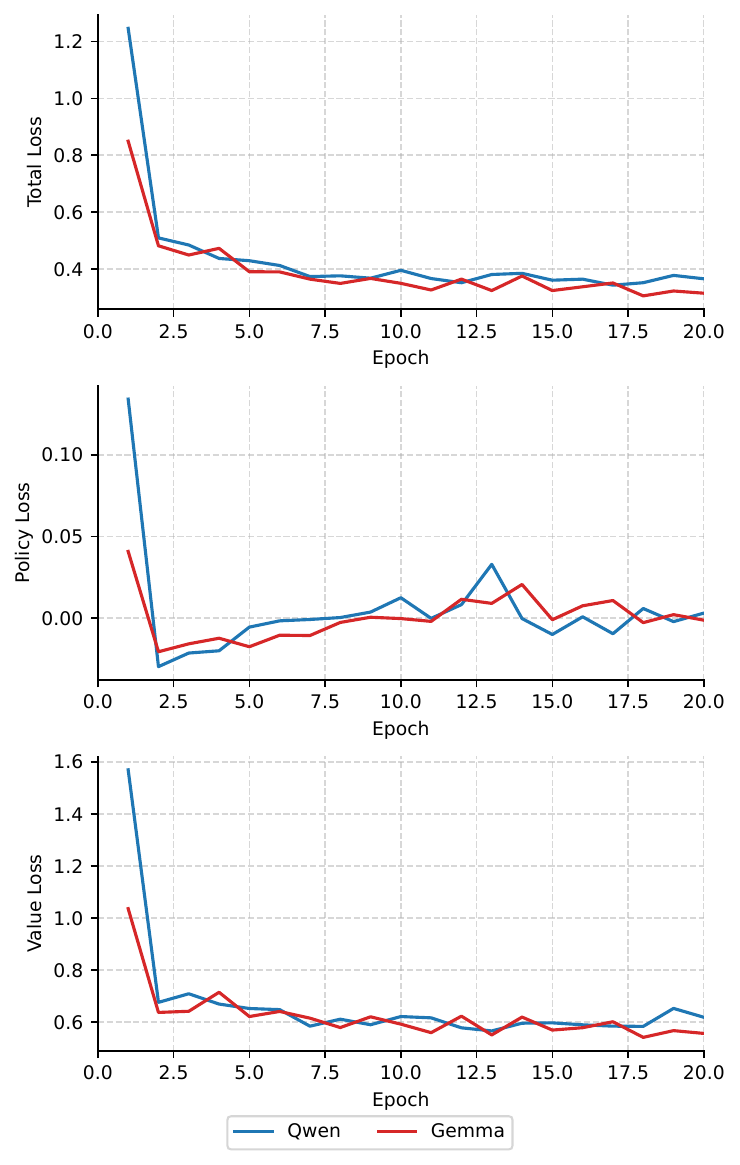}
    \caption{Representative training curves for Qwen 2.5–3B and Gemma 2–2B aggregated across datasets, demonstrating stable fine-tuning behavior under \textsc{BinaryPPO}.}
    \label{fig:loss_policy_curves}
\end{figure}

To investigate model learning dynamics, we track loss and policy metrics over training. \cref{fig:loss_policy_curves} presents representative curves for Qwen 2.5–3B and Gemma 2–2B, aggregated across multiple datasets, illustrating that both policy and value losses converge within the first two epochs. Fine-tuning with \textsc{BinaryPPO} demonstrates stable training, with negligible fluctuations in loss in later epochs.

\setcounter{figure}{5}
\begin{figure*}[b]
    \centering
    \includegraphics[width=1\linewidth]{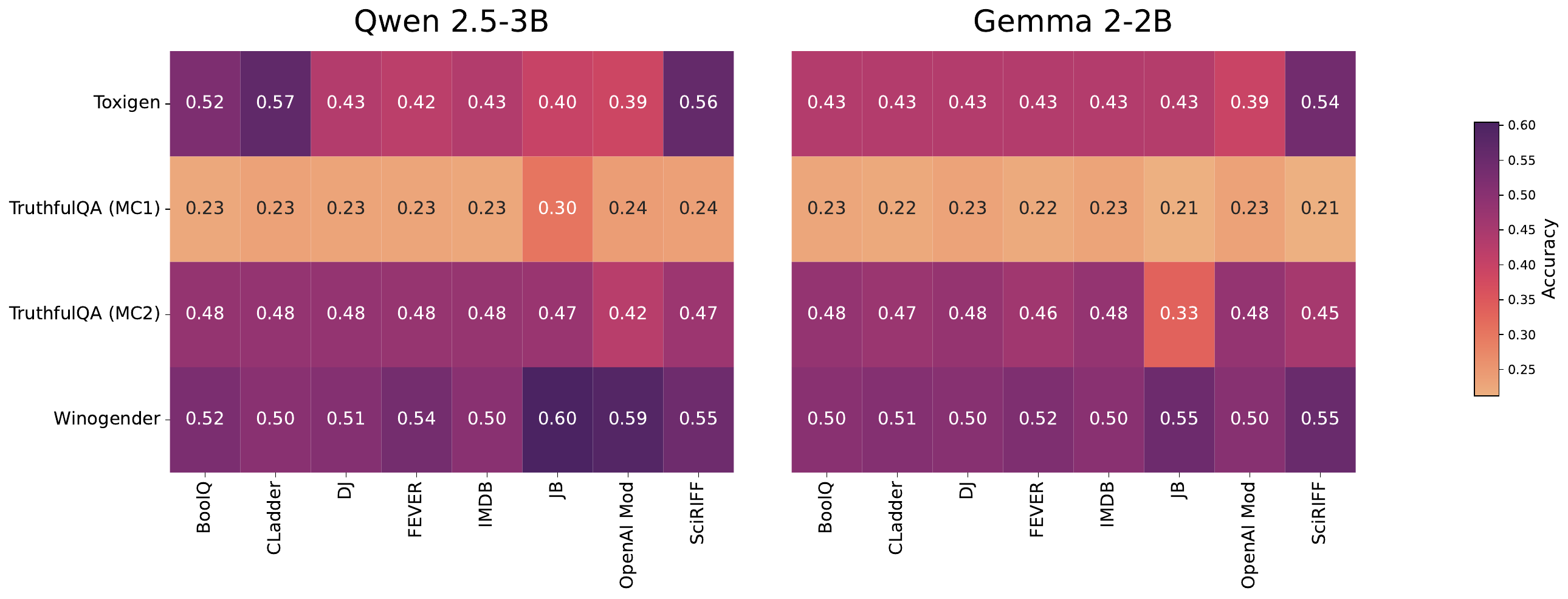}
    \caption{Normative evaluation of \textsc{BinaryPPO}-fine-tuned models across multiple benchmarks. Across datasets, \textsc{BinaryPPO} largely preserves or improves performance on toxicity detection (Toxigen), truthfulness assessment (TruthfulQA MC1/MC2), and gender bias evaluation (Winogender), with only limited regressions on select datasets. Overall, these results suggest that confidence-weighted reward shaping maintains core normative behaviors—truthfulness and fairness—while optimizing models for binary decision tasks, without inducing systematic degradation or catastrophic forgetting.}
    \label{fig:heatmap}
\end{figure*}

Additionally, we report evolutionary metrics of policy entropy and approximate KL divergence across training epochs in \cref{fig:other}. Entropy converges within the first seven epochs, with Qwen exhibiting higher entropy than Gemma. Similarly, KL divergence between successive policy distributions decreases rapidly, stabilizing within the first five epochs for both models.

\setcounter{figure}{4}
\begin{figure}[H]
    \centering
    \includegraphics[width=1\linewidth]{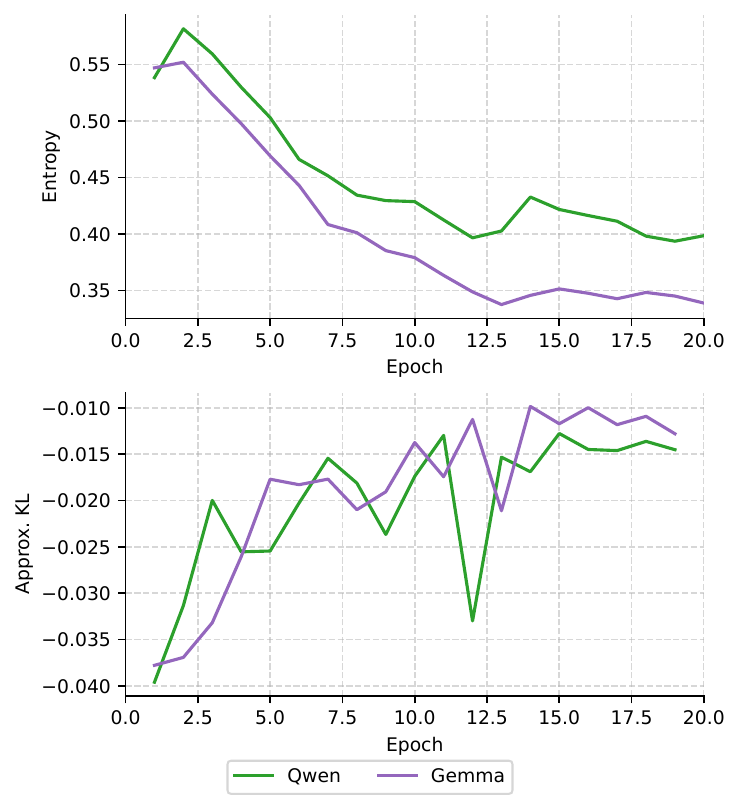}
    \caption{Evolution of policy metrics across epochs. (Top) Policy entropy converges within seven epochs, with Qwen maintaining higher entropy than Gemma. (Bottom) Approximate KL divergence between successive policies decreases and stabilizes within the first five epochs, indicating consistent policy refinement.}
    \label{fig:other}
\end{figure}

\subsection{Normative Evaluations}
\label{normative}

As a complementary axis, we assess whether fine-tuning with \textsc{BinaryPPO} preserves toxicity resistance, truthful selection, and gender neutrality across standard LLM benchmarks. Specifically, we evaluate the resulting models on a diverse set of tasks, including truthful question answering via TruthfulQA~\cite{lin2022truthfulqameasuringmodelsmimic}, toxicity assessment via Toxigen~\cite{hartvigsen2022toxigenlargescalemachinegenerateddataset}, and gender bias detection via Winogender~\cite{rudinger2018genderbiascoreferenceresolution}.

Overall, \textsc{BinaryPPO} maintains strong normative performance with limited and dataset-specific regressions. For toxicity, Qwen-based models exhibit mild degradations on BoolQ, CLadder, and SciRIFF, while Gemma-based models show noticeable drops only on SciRIFF, indicating improved robustness under \textsc{BinaryPPO} for Gemma. In contrast, gender bias evaluation via Winogender reveals consistent or improved accuracy for Qwen models,  when trained on the JB and OpenAI moderation datasets, while Gemma models largely preserve baseline performance across all training sources.

With respect to truthfulness, TruthfulQA MC1 accuracy remains comparatively low across all models. On the other hand, performance on TruthfulQA MC2 is consistently preserved, suggesting that \textsc{BinaryPPO} does not impair the model’s ability to identify truthful responses when evaluated under less adversarial multiple-choice formulations. Taken together, these results indicate that \textsc{BinaryPPO} does not induce systematic degradation of safety-critical or normative behaviors, and largely preserves truthfulness and fairness while optimizing for binary decision objectives.

\section{Discussion}

\subsection{Applications to AI Safety Detection}

Binary classification under noisy or ambiguous labels is critical in AI safety applications, including harmful content detection, model alignment evaluation, and moderation. \textsc{BinaryPPO}'s confidence-weighted reward approach enables models to make robust, calibrated decisions even in the presence of label noise or adversarial inputs. This framework can be applied to current harmfulness classifier pipelines where misclassifications carry high risk, providing a principled mechanism for LLMs to prioritize accurate, high-confidence decisions. Moreover, the offline nature of our method allows it to be deployed safely without exposing models to unsafe online environments involving misaligned agents, aiding in building robust classifiers.

\subsection{Sociopolitical Risks}

While \textsc{BinaryPPO} improves binary decision-making, caution is required when applying it to socially sensitive tasks such as assessing voter manipulation, polarization, or levels of radicalization from classifier datasets. Biases in training data or skewed reward signals could amplify existing societal biases or lead to overconfident judgments in ambiguous cases. As \textsc{BinaryPPO} demonstrates near-saturated performance upon dataset fine-tuning, we highlight the importance of careful reward design, and external model bias evaluations to mitigate unintended consequences. Future work should investigate fairness-aware reward shaping and mechanisms to ensure responsible deployment in socially impactful domains.

\subsection{Confidence in Reinforcement Learning}

A key insight from our work is the role of confidence in guiding reinforcement learning for language models. By incorporating probabilistic reward signals that scale with model confidence, \textsc{BinaryPPO} aligns learning objectives with decision quality rather than raw label agreement. Our experiments demonstrate that this approach stabilizes policy learning, improves cross-domain generalization, and enables better handling of minority classes and noisy labels. More broadly, confidence-weighted rewards may provide future strategy for applying RL to other discrete decision tasks in NLP, bridging the gap between probabilistic modeling and practical, high-stakes applications.

\section{Conclusion}

In this work, we introduced \textsc{BinaryPPO}, an offline reinforcement learning framework that reframes binary classification for large language models as a confidence-aware decision-making problem. By using a probabilistic reward signal that explicitly incorporates model confidence, \textsc{BinaryPPO} addresses several core limitations of supervised fine-tuning in settings characterized by noisy labels, class imbalance, and ambiguous supervision. Across a diverse suite of binary classification benchmarks spanning reasoning, safety, and factuality domains, our approach consistently outperforms strong SFT and PPO baselines, achieving substantial gains in accuracy while exhibiting stable training dynamics and out-of-distribution generalization.

More broadly, this work suggests that many binary NLP tasks traditionally approached through supervised learning may be better modeled as sequential decision problems under uncertainty. Offline reinforcement learning provides a principled framework for incorporating uncertainty, confidence calibration, and asymmetric error costs—properties that are difficult to encode in static loss functions. While this study focuses on binary classification, the underlying ideas extend naturally to multi-class and structured decision problems, as well as to settings where labels are weak, subjective, or partially observed. We view \textsc{BinaryPPO} as a step toward more robust, interpretable, and safety-aligned training paradigms for language models, and hope this work encourages further exploration of reinforcement learning as a practical alternative to supervised fine-tuning in high-stakes NLP applications.

\section*{Limitations}

This work focuses on binary decision tasks with discrete action spaces and evaluates \textsc{BinaryPPO} on small-to mid-sized open-weight LLMs due to compute constraints. While the proposed confidence-weighted reward formulation is general, extending it to multi-class and larger frontier-scale models remains future work. Additionally, optimal reward shaping and entropy regularization may require modest task-specific tuning, and our normative evaluations provide only partial coverage of real-world safety concerns.

\section*{Impact Statement}

Our work improves LLM robustness and calibration on tasks like toxicity detection, factuality verification, and causal reasoning, enabling safer content moderation, more reliable fact-checking, and stable deployment in high-stakes settings. By aligning model confidence with decision quality, \textsc{BinaryPPO} reduces misclassifications that could cause harm or propagate misinformation. We also note potential risks: overconfident predictions could amplify biases or cause harm in socially sensitive applications. Responsible deployment, fairness-aware reward design, and external evaluation are necessary to mitigate these risks. Overall, \textsc{BinaryPPO} provides a framework for more interpretable, robust, and safety-conscious AI systems.

\section*{Acknowledgement}
We thank the feedback on paper writing from Yongjin Yang and Jiarui Liu. This material is based in part upon work supported by the German Federal Ministry of Education and Research (BMBF): Tübingen AI Center, FKZ: 01IS18039B; by the Machine Learning Cluster of Excellence, EXC number 2064/1 – Project number 390727645; by Schmidt Sciences SAFE-AI Grant; by the Frontier Model Forum and AI Safety Fund; by Open Philanthropy; by the Survival and Flourishing Fund;
the University of Toronto’s Acceleration Consortium, which receives funding from the Canada First Research Excellence Fund (CFREF); and by the Cooperative AI Foundation. The usage of OpenAI credits is largely supported by the Tübingen AI Center and Schmidt Sciences. Resources used in preparing this research project were provided, in part, by the Province of Ontario, the Government of Canada through CIFAR, and companies sponsoring the Vector Institute.

\nocite{langley00}

\bibliography{main}
\bibliographystyle{icml2026}

\newpage
\appendix
\onecolumn
\section{Dataset Description}

This section provides detailed descriptions of all datasets used in the
\textsc{BinaryPPO} experimental pipeline. We additionally describe the
reconstruction procedure used to obtain an out-of-domain validation split
of the CLadder dataset for the experiments reported in
Section~\cref{ood}.

\subsection{Individual Datasets}
\label{ds_description}

All datasets used in this work are publicly available and framed as binary
classification tasks. Unless otherwise noted, we use the standard train-test
splits provided by the dataset authors and apply only minimal preprocessing to
preserve the original task semantics. Below, we briefly describe each dataset.

\begin{itemize}
    \item \textbf{CLadder}: CLadder~\cite{jin2024cladderassessingcausalreasoning} is a causal reasoning benchmark derived from explicit causal graphs and formal causal queries. Each example is generated by querying an oracle causal inference engine with associational, interventional, or counterfactual questions, and translating the resulting symbolic query into natural language.
    
    \item \textbf{BoolQ}: BoolQ~\cite{clark2019boolqexploringsurprisingdifficulty} is a reading comprehension dataset of naturally occurring yes/no questions paired with short Wikipedia passages. Questions are collected from unconstrained, real-world queries and often require non-trivial, entailment-like reasoning over the passage rather than direct fact lookup. The task is to determine whether the passage supports an affirmative answer to the question.
    
    \item \textbf{FEVER}: FEVER~\cite{thorne-etal-2018-fever} is a fact verification dataset with claims derived from Wikipedia and annotated as Supported, Refuted, or NotEnoughInfo. For our binary experiments, only Supported/Refuted claims are used, with evidence sentences provided for verification.

    \item \textbf{IMDB}: IMDB~\cite{maas-EtAl:2011:ACL-HLT2011} is a binary sentiment classification dataset of movie reviews labeled positive or negative. It is widely used to evaluate models’ ability to detect subjective sentiment in natural language.
    
    \item \textbf{JailbreakBench}: JailbreakBench~\cite{chao2024jailbreakbench} is a benchmark of adversarial prompts designed to elicit harmful or unsafe behavior from LLMs. Each example is labeled according to whether it successfully bypasses model safety, enabling standardized evaluation of jailbreak attacks.
    
    \item \textbf{OpenAI Moderation}: The OpenAI Moderation Dataset~\cite{openai2022moderation} is a dataset for content moderation, with examples labeled according to the presence of unsafe or undesired content (e.g., hateful, violent, or harassing text). It enables binary classification of text as safe or unsafe.
    
    \item \textbf{Detect-Jailbreak}: Detect-Jailbreak~\cite{SCBSZ24, zou2023universal} is a dataset of prompts designed to bypass LLM safety measures. Each example is labeled to indicate whether it successfully induces unsafe or disallowed model behavior, supporting binary safety classification. 
    
    \item \textbf{SciRIFF}: SciRIFF~\cite{wadden2025sciriffresourceenhancelanguage} is a dataset of expert-written, instruction-following instances for scientific literature tasks, including information extraction, summarization, question answering, claim verification, and classification. Each example pairs a detailed instruction with a research-context input and structured output, supporting binary or multi-class evaluation depending on the task.
\end{itemize}

\subsection{Reconstruction of CLadder}
\label{cladder}
To construct an out-of-domain validation set, we independently reconstructed a subset of approximately 17k examples using code developed by the one of the original authors of CLadder. This reconstruction was performed without access to the original training or test splits used in our experiments.

We explicitly verified that no examples from the original CLadder test set were
included in the reconstructed split, ensuring the absence of test-set contamination. The reconstructed dataset preserves the original task format and
label semantics, but differs in surface form and instance composition, allowing
it to serve as a controlled distribution shift for OOD evaluation.

\section{Runtime Environment Details}

All experiments were repeated multiple times to ensure that reported trends
are robust. Training was performed on a GPU workstation, and results were
monitored to confirm stability and convergence.

\subsection{Models}

Two pretrained language models with distinct architectures were used:
Qwen 2.5-3B and Gemma 2-2B. Using models with different
architectures allowed us to verify that observed behaviors were not specific
to a particular design or implementation.

\subsection{Hardware and Runtime Environment}

Experiments were run on a workstation equipped with an NVIDIA H100 GPU with
CUDA support. The software environment included Python 3.10 and key packages
such as \texttt{transformers}, \texttt{torch} and \texttt{numpy}.

\section{Normative Evaluations}

To assess potential unintended side effects of \textsc{BinaryPPO}, we
evaluated models post-training on benchmarks measuring truthfulness, gender
bias, and toxicity, using the \texttt{lm-eval} library to standardize
execution across models~\cite{eval-harness}.

The benchmarks used include TruthfulQA~\cite{lin2022truthfulqameasuringmodelsmimic}, measuring factual accuracy and resistance to generating false but human-plausible answers, using multiple-choice questions Toxigen~\cite{hartvigsen2022toxigenlargescalemachinegenerateddataset}, focusing on the tendency of models to produce toxic content in response to benign prompts. As well as Winogender~\cite{rudinger2018genderbiascoreferenceresolution}, which assesses gender bias by measuring whether pronouns are associated with gender-stereotypical professions.

\begin{table}[H]
    \centering
    \small
    \renewcommand{\arraystretch}{1.1}
    \begin{tabular}{lcccc}
        \toprule
        \textbf{\textsc{Dataset}} & \textbf{\textsc{Toxigen}} & \textbf{\textsc{TruthfulQA MC1}} & \textbf{\textsc{TruthfulQA MC2}} & \textbf{\textsc{Winogender}} \\
        \midrule
        BoolQ      & 0.5213 & 0.2252 & 0.4826 & 0.5236 \\
        CLadder    & 0.5681 & 0.2338 & 0.4839 & 0.5014 \\
        DJ         & 0.4319 & 0.2313 & 0.4838 & 0.5097 \\
        FEVER      & 0.4202 & 0.2289 & 0.4798 & 0.5361 \\
        IMDB       & 0.4330 & 0.2264 & 0.4796 & 0.5000 \\
        JBB        & 0.4000 & 0.3048 & 0.4741 & 0.6042 \\
        OpenAI     & 0.3883 & 0.2424 & 0.4234 & 0.5875 \\
        SciRIFF    & 0.5628 & 0.2375 & 0.4705 & 0.5458 \\
        \bottomrule
    \end{tabular}
    \caption{Normative evaluation results of Qwen-2.5-3B across toxicity, truthfulness, and gender bias benchmarks.}
    \label{tab:normative_qwen}
\end{table}
\vspace{-2em}

\begin{table}[H]
    \centering
    \small
    \renewcommand{\arraystretch}{1.1}
    \begin{tabular}{lcccc}
        \toprule
        \textbf{\textsc{Dataset}} & \textbf{\textsc{Toxigen}} & \textbf{\textsc{TruthfulQA MC1}} & \textbf{\textsc{TruthfulQA MC2}} & \textbf{\textsc{Winogender}} \\
        \midrule
        BoolQ      & 0.4319 & 0.2277 & 0.4833 & 0.5000 \\
        CLadder    & 0.4319 & 0.2240 & 0.4736 & 0.5097 \\
        DJ         & 0.4319 & 0.2326 & 0.4814 & 0.5042 \\
        FEVER      & 0.4319 & 0.2215 & 0.4639 & 0.5194 \\
        IMDB       & 0.4319 & 0.2313 & 0.4838 & 0.5014 \\
        JBB        & 0.4309 & 0.2130 & 0.3331 & 0.5472 \\
        OpenAI     & 0.3947 & 0.2338 & 0.4844 & 0.5042 \\
        SciRIFF    & 0.5394 & 0.2118 & 0.4528 & 0.5528 \\
        \bottomrule
    \end{tabular}
    \caption{Normative evaluation results of Gemma-2-2B across toxicity, truthfulness, and gender bias benchmarks.}
    \label{tab:normative_gemma}
\end{table}

\section{Additional Evaluation Metrics}

We report threshold-free metrics for binary classification to complement accuracy. AUROC measures the model’s ability to rank positive instances above negatives across all decision thresholds, while AUPRC captures precision–recall tradeoffs and is particularly informative under class imbalance.

\begin{table*}[h]
\centering
\small
\setlength{\tabcolsep}{6pt}
\resizebox{\textwidth}{!}{%
\begin{tabular}{l cccccccc}
\toprule
\addlinespace[0.3em]
\textbf{\textsc{Model}} &
\textbf{\textsc{Cladder}} &
\textbf{\textsc{SciRIFF}} &
\textbf{\textsc{BoolQ}} &
\textbf{\textsc{Fever}} &
\textbf{\textsc{IMDB}} &
\textbf{\textsc{Mod}} &
\textbf{\textsc{JB}} &
\textbf{\textsc{Detect}} \\
\midrule
\addlinespace[0.3em]
\textbf{Qwen 2.5-3B} 
& 99.46\% (99.47\%)
& 99.74\% (99.88\%)
& 99.63\% (99.77\%)
& 99.58\% (99.60\%)
& 91.00\% (91.36\%)
& 99.95\% (99.98\%)
& 99.97\% (99.97\%)
& 99.83\% (98.34\%) \\
\textbf{Gemma 2-2B}
& 99.46\% (99.47\%)
& 99.98\% (99.98\%)
& 99.89\% (99.91\%)
& 99.60\% (99.52\%)
& 99.91\% (99.92\%)
& 99.97\% (99.99\%)
& 99.97\% (99.97\%)
& 99.91\% (98.79\%) \\
\bottomrule
\end{tabular}
}
\vspace{0.5em}
\caption{Threshold-free evaluation results for Qwen 2.5-3B and Gemma 2-2B across all benchmarks. Values are reported as AUROC (AUPRC) in percentage form, confirming improved classification performance.}
\label{tab:auroc_combined}
\end{table*}


\end{document}